\documentclass[11pt]{article}
\usepackage{acl2012}
\usepackage{times}
\usepackage{latexsym}
\usepackage{amsmath}
\usepackage{multirow}
\usepackage{url}
\usepackage{color}
\usepackage{verbatim}
\usepackage{graphicx}
\usepackage[table]{xcolor}
\usepackage{booktabs}
\usepackage{subfig}
\definecolor{gris}{HTML}{DEDEDE}
\definecolor{white}{HTML}{FFFFFF}

\title{Identification of Fertile Translations in Medical Comparable Corpora :\\
a  Morpho-Compositional Approach}

\author{Estelle Delpech, B\'eatrice Daille, Emmanuel Morin \\
     Universit\'e de Nantes - LINA FRE CNRS 2729\\
 2 rue de la Houssini\`ere BP 92208 \\
 44322 Nantes Cedex 3, France\\
  {\tt \small \{name.surname\}@univ-nantes.fr}
 \\
 \And
 Claire Lemaire  \\
       Lingua et Machina \\
 c/o Inria Rocquencourt BP 105 \\
Le Chesnay Cedex 78153, France \\
  {\tt \small cl@lingua-et-machina.com} 
}

\date{}

\begin{document}
\maketitle
\begin{abstract}
This paper defines a method for lexicon in the biomedical domain from comparable corpora. The method is based on compositional translation and exploits morpheme-level translation equivalences. It can generate translations for a large variety of morphologically constructed words and can also generate 'fertile' translations. We show that fertile translations increase the overall quality of the extracted lexicon for English to French translation.
\end{abstract}

\section{Introduction}
Comparable corpora are composed of texts in different languages which are not translations but deal with the same subject matter and were produced in similar situations of communication so that there is  a possibility to find translation pairs in the texts. 
Comparable corpora have been used mainly in the field of Cross-Language Information Retrieval 
 and Computer-Aided Translation (CAT). In CAT, which is our field of application, comparable corpora have been used to extract domain-specific bilingual lexicons for language pairs or subject domains for which no parallel corpora is available.  Another advantage of comparable corpora is that they contain more idiomatic expressions than parallel corpora do. Indeed, the target texts of parallel corpora are translations and bear the influence of the source language whereas the target texts of comparable corpora are original, spontaneous productions.
 The main drawback of comparable corpora is that much fewer translation pairs can be extracted than in parallel corpora because (i) not all source language terms do have a translation in the target texts and (ii) when there is a translation, it may not be present in its canonical form, precisely because the target texts are not translations. As observed by Baker (1996)\nocite{baker_corpusbased_1996}, translated texts tend to bear features like explication, simplification, normalization and leveling out. For instance, an English-French comparable corpus may contain the English term \emph{post-menopausal} but not its ``normalized'' or ``canonical'' translation in French (\emph{post-m\'enopausique}). However, there might be some morphological or paraphrastic variants in the French texts like \emph{post-m\'enopause 'post-menopause'} or \emph{apr\`es la m\'enopause 'after the menopause'}.
The solution that consists in increasing the size of the corpus in order to find more translation pairs or to extract parallel segments of text  \cite{fung_mining_2004,rauf_use_2009} is only possible when large amounts of texts are available. In the case of the extraction of \emph{domain-specific} lexicons, we quickly face the problem of data scarcity: in order to extract high-quality lexicons, the corpus must contain text dealing with very specific subject domains and the target and source texts must be highly comparable. If one tries to increase the size of the corpus, one takes the risk of decreasing its quality by lowering its comparability or adding out-of-domain texts. Studies support the idea that the quality of the corpora is more important than its size. Morin et al. (2007) \nocite{morin_bilingual_2007} show that the discourse categorization of the documents increases the precision of the lexicon despite the data sparsity. Bo and Gaussier (2010) \nocite{bo_improving_2010} show that they improve the quality of a lexicon if they improve the comparability of the corpus by selecting a smaller - but more comparable - corpus  from an initial set of documents.
Consequently, one solution for increasing the number or translation pairs is to focus on identifying translation variants. This paper explores the feasibility of identifying "\emph{fertile}" translations in comparable corpora. In parallel texts processing, the notion of fertility has been defined by Brown et al. (1993)\nocite{brown_mathematics_1993}. They defined the fertility of a source word $e$ as the number of target words to which $e$ is connected in a randomly selected alignment. Similarly, we call a fertile translation a translation pair in which the target term has more words than the source term. 
We propose to identify such translations with a method mixing  morphological analysis and compositional translation : (i) the source term is decomposed into morphemes:   \emph{post-menopausal} is split into  \emph{post- + menopause}\footnote{We use the following notations for morphemes: trailing hyphen for prefixes (\emph{a-}), leading hyphen for suffixes (\emph{-a}), both for confixes (\emph{-a-}) and no hyphen for autonomous morphemes (\emph{a}). Morpheme boundaries are represented by a plus sign (\emph{+}).}  ; (ii) the morphemes are translated as bound morphemes or fully autonomous words:  \emph{post-} becomes \emph{post-} or \emph{apr\`es} and  \emph{menopause} becomes \emph{m\'enopause}  and (iii) the translated elements are recomposed into a target term: \emph{post-m\'enopause},  \emph{apr\`es la m\'enopause}.

This paper falls into 4 sections. 
Section \ref{sec:related} outlines recent research in compositionality-based lexicon extraction. 
Section \ref{sec:mct} explains the algorithm of morpho-compositional translation. Experimental data and results and described in sections \ref{sec:data}  and \ref{sec:evaluation}.

\section{Related work}
\label{sec:related}
Most of the research work in lexicon extraction from comparable corpora concentrates on same-length term alignment. To our knowledge, only Daille and Morin (2005)\nocite{daille_frenchenglish_2005} and Weller et al. (2011) \nocite{weller_simple_2011} tried to align terms of different lengths. Daille and Morin (2005)\nocite{daille_frenchenglish_2005} focus on the specific case of multi-word terms  whose meanings are non-compositional and tried to align these multi-word terms with either single-word terms or multi-word terms using a context-based approach\footnote{Context-based methods were introduced by Rapp (1995) and Fung (1997)\nocite{rapp_identifying_1995, fung_finding_1997}. They consist in comparing the contexts in which the source and target terms occur. Their drawback is that they need the source and target terms to be very frequent.}.
 Weller et al. (2011)\nocite{weller_simple_2011} concentrate on aligning German \textsc{Noun-Noun} compounds to \textsc{Noun Noun} and \textsc{Noun Prep Noun} structures in French and English.

We chose to work in the framework of compositionality-based translation because:
(i)  compositional terms form more than 60\% of the new terms found in techno-scientific domains, and especially in the field of biomedecine \cite{namer_defining_2007}
(ii) compositionality-based methods have been  shown to clearly outperform context-based ones for the translation of terms with compositional meaning \cite{morin_compositionality_2010} 
(iii) we believe that compositionality-based methods offer the opportunity to generate fertile translations if combined with a morphology-based approach.

\subsection{Principle of compositional translation}
\label{sec:compo_approach}
Compositional translation relies on the principle of compositionality which states that 
``the meaning of the whole is a function of the meaning of the parts'' \cite[24-25]{keenan_boolean_1985}.  Applied to bilingual lexicon extraction, compositional translation ($\mathcal{CT}$) consists in decomposing the source term into atomic components ($\mathcal{D}$), translating these components into the target language ($\mathcal{T}$), recomposing the translated components into target terms ($\mathcal{R}$) and finally filtering the generated translations with a selection function ($\mathcal{S}$):
{\center \small
\begin{quote}
$\mathcal{CT}\text{(``ab'')}$\\
$~=\mathcal{S(R(T(D(}\text{``ab''}))))$\\
$~=\mathcal{S(R(T(} \text{\{a, b\}})))$\\
$~=\mathcal{S(R(}\{\mathcal{T}\text{(a)}\times\mathcal{T}\text{(b)}\}))$\\
$~=\mathcal{S(R(}\text{\{\textsc{a, b}\}}))$\\
$~=\mathcal{S(}\text{\{\textsc{a, b}\}, \{\textsc{b, a}\}})$\\
$~=\text{``\textsc{ba}''}$\\
$~$ \\
\end{quote}
}
where ``ab'' is a source term composed of \emph{a} and \emph{b}, ``$\textsc{ba}$'' is a target term composed of $\textsc{b}$ and $\textsc{a}$ and there exists a bilingual resource linking \emph{a} to $\textsc{a}$ and \emph{b} to $\textsc{b}$. 

\subsection{Implementations of compositional translation}
\label{sec_implementations}
Existing implementations differ on the kind of atomic components they use for translation.

 {\bf Lexical compositional translation} \cite{grefenstette_world_1999,baldwin_translation_2004,robitaille_compiling_2006,morin_compositionality_2010} deals with multi-word term to multi-word term alignment  and uses lexical words\footnote{as opposed to grammatical words: preposition, determiners, etc.} as atomic components : \emph{rate of evaporation} is translated into French \emph{taux d'\'evaporation} by translating  \emph{rate} as \emph{taux} and \emph{evaporation} as \emph{\'evaporation} using  dictionary lookup. Recomposition may be done by permutating the translated components \cite{morin_compositionality_2010} or with translation patterns \cite{baldwin_translation_2004}. 

{\bf Sublexical compositional translation} deals with single-word term translation. The atomic components are subparts of the source single-word term. Cartoni (2009)\nocite{cartoni_lexical_2009} translates neologisms created by prefixation with a special formalism called Bilingual Lexeme Formation Rules. Atomic components are the prefix and the lexical base: Italian neologism \emph{anticonstituzionale 'anticonstitution'} is translated into French \emph{anticonstitution} by translating the prefix \emph{anti-} as \emph{anti-} and the lexical base \emph{constituzionale} as \emph{constitution}. 
Weller et al. (2011) \nocite{weller_simple_2011} translate two types of single-word term. German single-word term formed by the concatenation of two neoclassical roots are decomposed into these two roots, then the roots are translated into target language roots and recomposed into an English or French single-word term, e.g. \emph{Kalori$_1$metrie$_2$} is translated as \emph{calori$_1$metry$_2$}. German  \textsc{Noun$_1$-Noun$_2$} compounds are translated into French and English \textsc{Noun$_1 $Noun$_2$} or \textsc{Noun$_1$ Prep Noun$_2$} multi-word term, e.g. \emph{Elektronen$_{N1}$-mikroskop$_{N2}$} is translated as \emph{electron$_{N1}$ microscope$_{N2}$}.

\subsection{Challenges of compositional translation}
\label{sec_challenges}
Compositional translation faces four main challenges which are 
{\bf (i) morphosyntactic variation:} source and target terms' morphosyntactic structures are different: 
\emph{anti-\underline{cancer}$_{\textsc{Noun}}$ $\rightarrow$ anti-\underline{canc\'ereux}$_{\textsc{Adj}}$ 'anti-cancerous'} ;
{\bf (ii) lexical variation:} source and target terms contain semantically related - but not equivalent - words: \emph{\underline{machine} translation $\rightarrow$ traduction \underline{automatique} 'automatic translation'} ; 
{\bf (iii) fertility:} the target term has more content words than the source term: \emph{isothermal \underline{snowpack} $\rightarrow$ \underline{manteau} \underline{neigeux} isotherme 'isothermal snow mantel'} ;
{\bf (iv) terminological variation:} a source term can be translated as different target terms:
\emph{oophorectomy $\rightarrow$ ovariectomie 'oophorectomy', ablation des ovaires 'removal of the ovaries'}.

Solutions to morphosyntactic, lexical and to some extent terminological variation have been proposed in the form of thesaurus lookup \cite{robitaille_compiling_2006}, morphological derivation rules \cite{morin_compositionality_2010}, morphological variant dictionaries \cite{cartoni_lexical_2009} or morphosyntactic translation patterns \cite{baldwin_translation_2004,weller_simple_2011}. Fertility has been addressed by Weller et al. (2011)\nocite{weller_simple_2011} for the specific case of German \textsc{Noun-Noun} compounds.

\section{Morpho-compositional translation}
\label{sec:mct}
\subsection{Underlying assumptions}
Morpho-compositional translation (morpho-compositional translation) relies on the following assumptions:

{\bf Lexical subcompositionality.} 
The lexical items which compose a multi-word term or a single-word term may be split into semantically-atomic components. These components may be either
{\bf free} (i.e. they can occur in texts as autonomous lexical items  like \emph{toxicity} in \emph{cardiotoxicity}) or 
{\bf bound} (i.e. they cannot occur as autonomous lexical items, in that case they correspond to bound morphemes like \emph{-cardio-} in \emph{cardiotoxicity}).

{\bf Irrelevance of the bound/free feature in translation.}
Translation occurs regardless of the components' degree of freedom: \emph{-cardio-} may be translated as \emph{c\oe{}ur 'heart'} as in \emph{cardiotoxicity} $\rightarrow$ \emph{toxicit\'e pour le c\oe{}ur 'toxicity to the heart'}.

{\bf Irrelevance of the bound/free feature in allomorphy.}
Allomorphy happens regardless of the components' degree of freedom: \emph{-cardio-, c\oe{}ur 'heart', cardiaque 'cardiac'} are possible instantiations of the same abstract component and may lead to terminological variation as in \emph{cardiotoxicity} $\rightarrow$ \emph{cardiotoxicit\'e 'cardiotoxicity', toxicit\'e pour le c\oe{}ur 'toxicity to the heart', toxicit\'e cardiaque 'cardiac toxicity'}.

Like other sublexical approaches, the main idea behind  morpho-compositional translation is to go beyond the word level and work with subword components. In our case, these components are morpheme-like items which either (i) bear referential lexical meaning like confixes\footnote{we use the term \emph{confix} as a synonym of  neoclassical roots (Latin or Ancient Greek root words).} (\emph{-cyto-, -bio-, -ectomy-}) and autonomous lexical items (\emph{cancer, toxicity}) or (ii) can substantially change the meaning of a word, especially prefixes (\emph{anti-, post-, co-...}) and some suffixes (\emph{-less, -like...}). 
Unlike other approaches, morpho-compositional translation is not limited to small set of source-to-target structure equivalences. It takes as input a single morphologically constructed word unit which can be the result of prefixation '\emph{pretreatment}', confixation '\emph{densitometry}', suffixation '\emph{childless}', compounding '\emph{anastrozole-associated'} or any combinations of the four. It outputs a list of $n$ words who may or may not be morphologically constructed. For instance,  \emph{postoophorectomy} may be translated as \emph{postovariectomie 'postoophorectomy', apr\`es l'ovariectomie 'after the oophorectomy'} or \emph{apr\`es l'ablation des ovaires 'after the removal of the ovaries'}.

\subsection{Algorithm}
As an example, we show the translation of the  adjective \emph{cytotoxic} into French using a toy dataset.  Let $Comp^l_{type}$ be a list of components in language $l$ where $type$ equals $pref$ for prefixes, $conf$ for confixes, $suff$ for suffixes and $free$ for free lexical units ; $Trans$ be the translation table which maps source and target components ; $Var^l$ be a table mapping related lexical units in language $l$ ;  $Stop^l$ a list of stopwords in language $l$ ; $Corpus^l$ a lemmatized, pos-tagged corpus in language $l$:
{\small
\begin{tabular}{l}
$ Comp^{en}_{conf} = \{\text{-cyto-}\}$ ;\\
$ Comp^{en}_{free} =\{\text{cytotoxic, cytotoxicity, toxic}\}$ ;\\
$ Comp^{fr}_{conf} = \{\text{-cyto-}\}$ ;\\
$ Comp^{fr}_{free} = \{\text{cellule, toxique}\}$ ;\\
$Trans = \{\{\text{-cyto-} \rightarrow \text{-cyto-}, \text{cellule}\},$\\
\qquad\qquad$\{\text{toxic} \rightarrow \text{toxique}\}\}$ ;\\
$Var^{en} = \{\text{cytoxic} \rightarrow \text{cytoxicity}\}$ ;\\
$ Stop^{fr} = \{\text{pour}, \text{le}\}$ ;\\
$Corpus^{fr} = \text{``le/\textsc{det} cytotoxicit\'e/\textsc{n} \^etre/\textsc{aux} le/\textsc{det}}$\\
$\text{propri\'et\'e/\textsc{n} de/\textsc{prep} ce/\textsc{det} qui/\textsc{pro} \^etre/\textsc{aux}}$\\
$\text{toxique/\textsc{a} pour/\textsc{prep} le/\textsc{det} cellule/\textsc{n} ./\textsc{pun}''}$ ;\\
'\emph{The cytotoxicity is the property of what is toxic to}\\
\emph{the cells.}'
\end{tabular}}

~\\
Morpho-compositional translation  takes as input a source language single-word term and outputs zero or several target language single-word terms or multi-word terms. It is the result of the sequential application of four functions to the input single-word term: decomposition ($\mathcal{D}$), translation ($\mathcal{T}$), recomposition ($\mathcal{R}$) and selection ($\mathcal{S}$).
%

\subsubsection{Decomposition function}
The decomposition function $ \mathcal{D}$ works in two steps $ \mathcal{D}_1$ and $ \mathcal{D}_2$.

{\bf Step 1 of decomposition ($ \mathcal{D}_1$) } splits the input single-word term into minimal components by matching substrings of the single-word term with the resources $Comp^{src}$, $Comp^{src}_{conf}$, $Comp^{src}_{suff}$, $Comp^{src}_{free}$ and respecting some length constraints on the substrings. For example, one may split a single-word term $SWT_{1,n}$ of $n$ characters into prefix $Pref_{1,i}$ and lexical base  $LexBase_{i+1,n}$ provided that $SWT_{1,i} \in Comp^{src}_{pref}$ and $SWT_{i+1,n} \in Comp^{src}_{free}$ and $n-i > \mathcal{L}0 $ ; $ \mathcal{L}0 $ being empirically set to 5. A single-word term is first split into an optional prefixe + base$_1$, then base$_1$ is split into base$_2$ + optional suffix, then base$_2$ is split into one or several confixes or lexical items. When several splittings are possible, only the ones with the highest number of minimal components are retained. 
\begin{quote}\small
$\mathcal{S(R(T(D}_2\mathcal{(D}_1(\text{``cytotoxic'''})))))$\\
$~=\mathcal{S(R(T(D}_2( \text{\{cyto, toxic\}}))))$
\end{quote}

{\bf Step 2 of decomposition ($ \mathcal{D}_2$) } gives out all possible decompositions of the single-word term by enumerating the different concatenations of its minimal components. For example, if single-word term ``abc'' has been split into minimal components \{a,b,c\}, then it has 4 possible decompositions: \{abc\}, \{a,bc\}, \{ab,c\}, \{a,b,c\}. For a single-word term having $n$ minimal components, there exists $2^{n-1}$ possible decompositions.
\begin{quote}\small
$\mathcal{S(R(T(D}_2( \text{\{cyto, toxic\}}))))$\\
$~=\mathcal{S(R(T(} \text{\{cyto, toxic\}, \{cytotoxic\}})))$
\end{quote}

The concatenation of the minimal components into bigger components increases the chances of finding translations. For example, consider the single-word term \emph{non-cytotoxic} and a dictionary having translations for \emph{non}, \emph{cyto} and \emph{cytotoxic} but no translation for \emph{toxic}. If we stick to the sole output of $ \mathcal{D}_1$ \emph{\{non-,-cyto-,toxic\}}, the translation of \emph{non-cytotoxic} will fail because there is no translation for \emph{toxic}. Whereas if we also consider the output of $ \mathcal{D}_2$ which contains the decomposition \emph{\{non-,cytotoxic\}}, we will be able to translate \emph{non-cytotoxic} because the dictionary has an entry for \emph{cytotoxic}.

\subsubsection{Translation function}
The translation function $ \mathcal{T}$ provides translations for each decomposition output by $\mathcal{D}$. Applying the compositionality principle to translation, we consider that the translation of the whole is a function of the translation of the parts: $ \mathcal{T}(a,b) \cong \mathcal{T}(a) \times \mathcal{T}(b)$. For a given decomposition $\{c_1, ...c_n\}$ having $n$ components, there exists $ \prod^n_ {i=1}|\mathcal{T}(c_i)|$ possible translations. 
Components' translations are obtained using the $Trans$ and $Var$ resources: $\mathcal{T}(c) = Trans(c) ~\cup~ Trans(Var^{src}(c)) ~\cup ~Var^{tgt}(Trans(c))$. If one of the component cannot be translated, the translation of the whole decomposition fails. 

\begin{quote}\small
$\mathcal{S(R(T(} \text{\{cyto, toxic\}, \{cytotoxic\}})))$\\
$~=\mathcal{S(R(}\mathcal{T}(\text{cyto})\times\mathcal{T}(\text{toxic}),\mathcal{T}(\text{cytotoxic})))$\\
$~=\mathcal{S(R(}\text{\{cyto, toxique\},\{cellule, toxique\},}$\\
$~\text{ \{cytotoxicit\'e\}}))$\\
\end{quote}

\subsubsection{Recomposition function}
The recomposition function $\mathcal{R}$ takes as input the translations outputted by $\mathcal{T}$ and recomposes them into sequences of one or several lexical items. It takes place in two steps. 

{\bf Step 1 of recomposition ($\mathcal{R}_1$) } generates, for a given translation of $n$ items, all of the $n!$ possible permutations of these items. 
As a general rule, $O(n!)$ procedures should be avoided but we are permuting small sets (up to 4 items). This captures the fact that components' order may be different in the source and target language (distortion). 
Once the components have been permutated, we generate, for each permutation, all the different concatenations of its components into lexical items (like it is done in step 2 of decomposition). 
\begin{quote}\small
$\mathcal{S(R}_2\mathcal{(R}_1(\text{\{cyto,toxique\},\{cellule,toxique\},}$\\
$\text{ \{cytotoxicit\'e\}}))) $\\
$~=\mathcal{S(R}_2(\text{\{cyto,toxique\}, \{cytotoxique\},}$\\
$\text{\{toxique,cyto\}, \{toxiquecyto\}, \{cellule, toxique\},}$\\
$\text{\{celluletoxique\}, \{toxique, cellule\},}$\\ 
$\text{\{toxiquecellule\}, \{cytotoxicit\'e\}}))$
\end{quote}

{\bf Step 2 of recomposition ($\mathcal{R}_2$) } filters out the ouput of $\mathcal{R}_1$ using heuristic rules. For example, a sequence of lexical items $L = \{l_1, ...l_n\}$ would be filtered out provided that $ \exists ~ l \in L ~ | ~ l \in Comp^{tgt}_{pref} \cup Comp^{tgt}_{conf} \cup Comp^{tgt}_{suff}$, i.e. recomposition $\{\text{cytotoxique}\}$ would be accepted but not $\{\text{-cyto-}, \text{toxique}\}$ because \emph{-cyto-} is a bound component (it should not appear as an autonomous lexical item).

\begin{quote}\small
$\mathcal{S(R}_2(\text{\{cyto,toxique\}, \{cytotoxique\},}$\\
$\text{\{toxique,cyto\}, \{toxiquecyto\}, \{cellule, toxique\},}$\\
$\text{\{celluletoxique\}, \{toxique, cellule\},}$\\ 
$\text{\{toxiquecellule\}, \{cytotoxicit\'e\}}))$\\
$~=\mathcal{S(}\text{\{cytotoxique\}, \{toxiquecyto\},}$\\
$\text{\{cellule, toxique\}, \{celluletoxique\},}$\\ 
$\text{\{toxique, cellule\}, \{toxiquecellule\}, \{cytotoxicit\'e\}})$
\end{quote}

These concatenations correspond to the final lexical units which will be matched against the target corpus with the selection function. For example,  the concatenation $\{\text{toxique}_\textsc{a}, \text{cellule}_\textsc{b}\}$ corresponds to a translation made of two distinct lexical items: \emph{toxique} followed by \emph{cellule}. The concatenation $\{\text{cytotoxique}_\textsc{ab}\}$ corresponds to only one lexical item: \emph{cytotoxique}.

\subsubsection{Selection function}
The selection function $ \mathcal{S}$ tries to match the sequences of lexical items outputted by $ \mathcal{R}$ with the lemmas of the tokens of the target corpus.
We call $T = \{t_1, ...t_m\}$ a sequence of tokens from the target corpus, $l(t_k)$ the lemma of token $t_k$ and $p(t_k)$  the part-of-speech of token $t_k$.
We call $L = \{l_1, ...l_n\}$ a sequence of lexical items outputted by $ \mathcal{R}$. 
$L$ matches $T$ if 
there exists a strictly increasing sequence of indices $I = \{i_1, ...i_n\}$ 
such as $l(t_{i_{j}}) = l_j$ 
and $\forall j , 1 \leq j \leq n$
and $\forall i , 1 \leq |i_{j-1} - i_j| \leq \mathcal{L}1$ 
and 
$\forall t_k | ~ k \notin I, ~ l(t_k) \in Stop^{tgt} $ ; $\mathcal{L}1$  being empirically set to 3.
\begin{quote}\small
$ ~ =\mathcal{S(}\text{\{cytotoxique\}, \{toxiquecyto\},}$\\
$ \text{\{cellule, toxique\}, \{celluletoxique\},}$\\ 
$ \text{\{toxique, cellule\}, \{toxiquecellule\}, \{cytotoxicit\'e\}}) $\\
$ ~ =\text{``cytotoxicit\'e/\textsc{n}'', ``toxique/\textsc{a} pour/\textsc{prep} le/\textsc{det}}$\\
$ \text{cellule/\textsc{n}''}$\\
'\emph{cytotoxicity}', '\emph{toxic to the cells}'
\end{quote}
In other words, $L$ is a subsequence of the lemmas of $T$ and we allow at maximum $\mathcal{L}1$ closed-class words between two tokens which match the lemmas of $L$.

For a given sequence of lexical items $L$, we collect from the target corpus all sequences of tokens $T_1, T_2, ... T_p$ which match $L$ according to our above-mentioned definition. We consider two sequences $T1$ and $T2$ to be equivalent candidate translations if $ |T1| = |T2| $ and $\forall (t1_i, t2_j) $ such that $ t1 \in T1, t2 \in T2, i=j $ then $ l(t1_i) = l(t2_j) $ and $ p(t1_i) = p(t2_j) $, i.e. if two sequences of tokens correspond to the same sequence of (lemma, pos) pairs, these two sequences are considered as the same candidate translation.

\section{Experimental data}
\label{sec:data}
We worked with three languages: English as source language and French and German as target languages.

\subsection{Corpora}
Our corpus is composed of specialized texts from the medical domain dealing with breast cancer. We define specialized texts as texts being produced by domain experts and directed towards either an expert or a non-expert readership \cite{bowker_working_2002}. The texts were manually collected from scientific papers portals and from information websites targeted to breast cancer patients and their relatives. Each corpus has approximately 400k words (cf. table~\ref{table_corpus_size}). All the texts were pos-tagged and lemmatized using the linguistic analysis suite \textsc{Xelda}\footnote{\scriptsize\url{http://www.temis.com}}. We also computed the comparability of the corpora. We used the comparability measure defined by \cite{bo_improving_2010} which indicates, given a bilingual dictionary, the expectation of finding  for each source word of the source corpus its translation in the target corpus and \emph{vice-versa}. The English-French corpus' comparability is 0.71 and the English-German corpus' comparability is 0.45. The difference in comparability can be explained by the fact that German texts on breast cancer were hard to find (especially scientific papers): we had to collect texts in which breast cancer was not the main topic. 
\begin{table}[h]
\small \center
\rowcolors{1}{white}{gris}
\begin{tabular}{l|l|l|l}
\hline
Readership&EN&FR&DE\\ \hline
 \hline
experts&218.3k&267.2k&197.2k\\
non-experts &198.2k&184.5k&201.7k\\ 
\hline \hline
TOTAL &416.5k&451.75k&398.9k\\
\hline  \hline
\end{tabular}
\caption{Composition and size of corpora in words}
 \label{table_corpus_size}
\end{table}

\subsection{Source terms}
\label{sec_source_terms}
We tested our algorithm on a set of source terms extracted from the English texts. The extraction was done in a semi-supervised manner.
\textbf{Step 1:} We wrote a short seed list of English bound morphemes.  We automatically extracted from the English texts all the words that contained these morphemes. For example, we extracted the words \emph{postchemotherapy} and \emph{poster} because they contained the string \emph{post-} which corresponds to a bound morpheme of English.
\textbf{Step 2:} The extracted words were sorted : those which were not morphologically constructed were eliminated (like \emph{poster}), and those which were morphologically constructed were kept (like \emph{postchemotherapy}). The morphologically constructed words were manually split into morphemes. For example,  \emph{postchemotherapy} was split into \emph{post-, -chemo-} and \emph{therapy}.
\textbf{Step 3:} If some bound morphemes which were not in the initial seed list were found when we split the words during step 2, we started the whole process again, using the new bound morphemes to extract new morphologically constructed words.
We also added hyphenated terms like \emph{ER-positive} to our list of source terms. 

We obtained a set 2025 English terms with this procedure. For our experiments, we excluded from this set the source terms which had a translation in the general language dictionary and whose translation was present in the target texts. The final test set for English-to-French experiments contains 1839 morphologically constructed source terms. The test set for English-to-German contains 1824 source terms.

\subsection{Resources used in the translation step $\mathcal{T}$} 
\label{sec_translation_resources}
Tables~\ref{tab:ress_trans} and \ref{tab:ress_var} show the size of the resources we used for translation.

\textbf{General language dictionary} We used the general language dictionary which is part of the linguistic analysis suite  \textsc{Xelda}.

\textbf{Domain-specific dictionary} We built this resource automatically by extracting pairs of cognates from the comparable corpora. We used the same technique as \cite{hauer_clustering_2011}: a SVM classifier trained on examples taken from online dictionaries\footnote{\scriptsize \url{http: //www.dicts.info/uddl.php}}.

\textbf{Morpheme translation table} To our knowledge, there exists no publicly available morphology-based bilingual dictionary. 
Consequently, we  asked translators to create an \emph{ad hoc} morpheme translation table for our experiment. This morpheme translation table links the English bound morphemes contained in the source terms to their French or German equivalents. The equivalents can be bound morphemes or lexical items.

In order to handle the variation phenomena described in section~\ref{sec_challenges}, we used a \textbf{dictionary of  synonyms} and lists of  \textbf{morphologically related words}. The dictionary of synonyms is the one part  of the \textsc{Xelda} linguistic analyzer. The lists of morphologically related words were built by stemming the words of the comparable corpora and the entries of the bilingual dictionary with a simple stemming algorithm 
 \cite{porter_algorithm_1980}. 

\begin{table}[h]
\small \center
\rowcolors{1}{white}{gris}
\begin{tabular}{l|l|l}
\hline
& EN$\rightarrow$FR&EN$\rightarrow$DE \\ \hline
\hline 
General language&38k$\rightarrow$60k&38k$\rightarrow$70k\\ \hline  \hline
Domain-specific&6.7k$\rightarrow$6.7k&6.4k$\rightarrow$6.4k\\ \hline  \hline
Morphemes (\textsc{total})&242$\rightarrow$729&242$\rightarrow$761\\ \hline
$\qquad$prefixes&50$\rightarrow$134&50$\rightarrow$166\\ \hline
$\qquad$confixes&185$\rightarrow$574&185$\rightarrow$563\\ \hline
$\qquad$suffixes&7$\rightarrow$21&7$\rightarrow$32\\ \hline
\hline
\end{tabular}
\caption{Nb. of entries in the multilingual resources}
\label{tab:ress_trans}
\end{table}
\begin{table}[h]
\small \center
\rowcolors{1}{white}{gris}
\begin{tabular}{l|l|l|l}
\hline
& EN$\rightarrow$EN&FR$\rightarrow$FR&DE$\rightarrow$DE \\ \hline
\hline 
Synonyms&5.1k$\rightarrow$7.6k&2.4k$\rightarrow$3.2k&4.2k$\rightarrow$4.9k\\ \hline
Morphol.&5.9k$\rightarrow$15k&7.1k$\rightarrow$18k&7.4k$\rightarrow$16k\\ \hline
\hline
\end{tabular}
\caption{Nb. of entries in the monolingual resources}
\label{tab:ress_var}
\end{table}

\subsection{Resources used in the decomposition step ($\mathcal{D}$)}  
The decomposition function uses the entries of the bound morphemes translation table (242 entries) and a list of  85k lexical items composed of the entries of the general language dictionary and English words extracted from the Leipzig Corpus~\cite{quasthoff_corpus_2006} which is a general language corpus.

\section{Evaluation}
\label{sec:evaluation}

\subsection{Evaluation metrics}
As explained in section~\ref{sec_implementations}, compositional translation consists in \emph{generating} candidate translations. These candidate translations can be filtered out with a classifier \cite{baldwin_translation_2004}, by keeping only the translations which occur in the target texts of the corpus \cite{weller_simple_2011,morin_compositionality_2010} or by using a search engine \cite{robitaille_compiling_2006}.
Unlike alignment evaluation in parallel texts, there is no reference alignmens to which the selected translations can be compared and we cannot use standard evaluation metrics like AER \cite{och_comparison_2000}. It is also difficult to find reference lexicons in specific domains since the goal of the extraction process is to create such lexicons. Furthermore, we also wish to evaluate if the algorithm can identify non-canonical translations which, by definition, can not be found in a reference lexicon. Usually, the candidate translations are annotated manually as \emph{correct} or \emph{incorrect} by native speakers. Baldwin and Takana (2004) \nocite{baldwin_translation_2004} use two standards for evaluation: \emph{gold-standard}, \emph{silver-standard}. Gold-standard is the set of candidate translations which correspond to canonical, reference translations. Silver-standard corresponds to the gold-standard translations plus the translations which ``capture the basic semantics of the source language expression and allow the source language expression to be recovered with reasonable confidence'' (op. cit.). 

The first evaluation metric is the \emph{precision} $P$ which is the number of correct candidate translations $|Corr|$ over the total number of generated candidate translations $|A|$: $P = \frac{|Corr|}{|A|}$.
In addition to precision, we propose to indicate the \emph{coverage}  $C$ of the lexicon, i.e. the proportion of source terms ($\textsc{st}$) which obtained at least one candidate translation regardless of its accuracy:
\begin{equation*}
C = \frac{ \sum^{|\textsc{st}|}_{i=1} \alpha (\textsc{st}_i)}{|\textsc{st}|}
\end{equation*}
were $\alpha (\textsc{st}_i) $ returns $1$ if $|A(\textsc{st}_i)| \geq 1$ else $0$.
As augmenting coverage tends to lower precision, we also compute $OQ$, the \emph{overall quality} of the lexicon, to get  an idea of the coverage/precision tradeoff:  $OQ= P \times C$.

\subsection{Results}

Compositional-translation methods give better results when they are applied to general language texts rather than domain-specific texts. This is due to the fact that the translations of the components can be easily found in dictionaries since they belong to the general language and it is also easier to collect large corpora. \textbf{Working with general language texts}, Baldwin and Takana (2004)\nocite{baldwin_translation_2004} were able to generate candidate translations for 92\% of their source terms and they report 43\% (gold-standard) to 84\% (silver standard) of correct translations. The size of their corpus exceeds 80M words for each language. Cartoni (2009)\nocite{cartoni_lexical_2009} works on the translation of prefixed Italian neologisms into French. He considers that the generated neologisms have a ``confirmed existence'' if they occur more than five times on Internet. He finds that between 42\% and 94\% of the generated neologisms fall into that category.
\textbf{Regarding domain-specific translation}, Robitaille et al. (2006)\nocite{robitaille_compiling_2006} use a search engine to build corpus from the web and incrementally collect translation pairs. They start with a list of 9.6 pairs (on average) with a precision of 92\% and end up with a final output of 19.6 pairs on average with a precision of 81\%. Morin and Daille (2009)\nocite{morin_compositionality_2010} could generate candidate translations for 15\% of their source terms and they report 88\% of correct alignments. The size of their corpus is 700k words per language.   Weller et al. (2011)\nocite{weller_simple_2011} were able to generate 8\% of correct French translations and 18\% of correct English translations for their 2000 German compounds. Their corpus contains approximately 1.5M words per language.

We ran the morpho-compositional translation prototype on the set of  source terms described in section~\ref{sec_source_terms}. 
The output candidate translations were manually annotated by two translators. Like Baldwin and Takana (2004), we used three annotation values: canonical translation, recoverable translation and incorrect.  In our case, recoverable translations correspond paraphrastic and morphological translation variants. For example, the canonical translation for \emph{post-menauposal} is \emph{post-m\'enopausique}. Recoverable translations are \emph{post-m\'enopause 'post-menopause'} and \emph{apr\`es la m\'enopause 'after the menopause'}. Fertile translations can be canonical translations if a non-fertile translation would have been more awkward. For example, the canonical translation for \emph{oestrogen-sensitive} is \emph{sensible aux \oe{}strog\`enes 'sensitive to oestrogens'}. A non-fertile translation would sound very unnatural.
 We computed  inter-annotator agreement on a set of 100 randomly selected candidate translations. We used the Kappa statistics \cite{carletta_assessing_1996} and obtained a high agreement (0.77 for English to German translations and 0.71 for English to French).

First, we tested the impact of the linguistic resources described in section~\ref{sec_translation_resources}  (B for \underline{B}aseline dictionaries, D for \underline{D}omain-specific dictionary, S for \underline{S}ynonyms, M for \underline{M}orphologically related words).  We also tested a simple \underline{Pre}fix+lemma  translation (Pref) in similar vein to the work of Cartoni (2009) to serve as a line of comparison with our method.
The results are given in tables \ref{tab_scores_fr} and \ref{tab_scores_de}.
The best results in terms of overall quality are obtained with the combination of the baseline and domain-specific dictionaries (BD). Morphologically related words and synonyms increase coverage to the cost of precision. 
Regarding English-to-French translations, we were able the generate translations for 26\% of the source terms. The gold-standard precision is 60\% and the silver standard precision is 67\%.
Regarding English-to-German translations, we were able the generate translations for 26\% of the source terms. The gold-standard precision is 39\% and the silver-standard precision is 43\%. 
The prefix+lemma translation method has a very high precision (between 84\% and 76\%) but produces very few translations (between 1\% and 2\%).
Coverage and precision scores compare well with other approaches knowing that we have very small domain-specific corpora (400k words per language) and that our approach deals with a large number of morphological constructions. The lower quality of the German translations can be explained by the fact that the English-German corpus is much less comparable than the English-French corpus (0.45 vs. 0.71).

\begin{table}[h]
\small \center
\rowcolors{1}{white}{gris}
\begin{tabular}{l|c|cc|cc} 
\hline
	&C	&\multicolumn{2}{c|}{P}&\multicolumn{2}{c}{OQ}\\
	&	&\textsc{gold}	&\textsc{silver}	&\textsc{gold}	&\textsc{silver}\\
\hline \hline
Pref &.01&.84&.9&.01&.01\\
\hline \hline
B&.12&.57&.60&.07&.07\\
BS&.15&.50&.53&.08&.08\\
BM&.23&.28&.37&.06&.09\\
BD&.26&\bf{.60}&\bf{.67}&\bf{.16}&\bf{.17}\\
BSMD&\bf{.39}&.33&.44&.13&\bf{.17}\\
\hline \hline	
\end{tabular}
\caption{Scores for the EN$\rightarrow$FR lexicon}
\label{tab_scores_fr}
\end{table}
\begin{table}[h]
\small \center
\rowcolors{1}{white}{gris}
\begin{tabular}{l|c|cc|cc} 
\hline
	&C	&\multicolumn{2}{c|}{P}&\multicolumn{2}{c}{OQ}\\
	&	&\textsc{gold}	&\textsc{silver}	&\textsc{gold}	&\textsc{silver}\\
\hline \hline
Pref&.02&.76&.86&.02&.02\\
\hline \hline
B&.13&.35&.39&.05&.05\\
BS&.16&.31&.35&.05&.05\\
BM&.22&.23&.29&.05&.06\\
BC&.26&\bf{.39}&\bf{.43}&{\bf.10}&.11\\
BCSM&\bf{.36}&.27&.34&{\bf.10}&\bf{.12}\\
\hline \hline	
\end{tabular}
\caption{Scores for the EN$\rightarrow$DE lexicon}
\label{tab_scores_de}
\end{table}

We also tested the impact of the fertile translations on the quality of the lexicon. Tables ~\ref{tab_diff_scores_fr} and ~\ref{tab_diff_scores_de} show the evaluation scores with and without  fertile translations. As expected, fertile translations enables us to increase the size of the lexicon but they are less accurate than non-fertile translations. Fertile translations increase the overall quality of the English-French lexicon by 4\% to 5\%. This is not the case for English-German translations: fertile translations result in a big drop in precision. The overall quality does not significantly change. This might  be partly due to the low comparability of the corpus but we think that the main reason lies in the morphological type of the languages involved in the translation. It is worth noticing that, if we consider only the non-fertile translations, the English-German lexicon has generally better scores than the English-French one.
In fact, fertile variants are more natural and frequent in French than in German. English and German are Germanic languages with a tendency to build new words by agglutinating words or morphemes into one single word. Noun compounds such as \emph{oestrogen-independent} or \emph{\"Ostrogen-unabh\"angige} are common in these two languages. Conversely, French is a Romance language which prefers to use phrases composed of two nouns and a preposition rather than a single-noun compound (\emph{oestrogen-independent}  would be translated as \emph{ind\'ependant des \oe{}strog\`enes 'independent to oestrogens'}).  It is the same with the bound morpheme/single word alternance. The term \emph{cytoprotection} will be translated into German as \emph{Zellschutz} whereas in French it can be translated as \emph{cytoprotection} or \emph{protection de la cellule 'protection of the cell'}.
\begin{table}[h]
\small \center
\rowcolors{1}{white}{gris}
\begin{tabular}{l|ll|ll|ll} 
\hline
&\multicolumn{2}{c|}{C}&\multicolumn{2}{c|}{P}&\multicolumn{2}{c}{OQ}\\
\hline \hline		
&-f&+f&-f&+f&-f&+f\\
\hline	 \hline	
\footnotesize B&.04&.12&{\bf.81}&.57&.03&.07\\
\footnotesize BS&.05&.15&.69&.50&.03&.08\\
\footnotesize BM&.11&.23&.20&.28&.02&.06\\
\footnotesize BD&.16&.26&.70&{\bf.60}&{\bf.11}&{\bf.16}\\
\footnotesize BSMD&{\bf.24}&{\bf.39}&.31&.33&.07&.13\\
\hline \hline	
avg. gain&\multicolumn{2}{c|}{\cellcolor[gray]{0.9}+11}&\multicolumn{2}{c|}{\cellcolor[gray]{0.9}-8.6}&\multicolumn{2}{c|}{\cellcolor[gray]{0.9}+4.8}\\
\hline \hline		
\end{tabular}
\caption{Scores without (-f) and with (+f) fertile translations (EN$\rightarrow$FR)}
\label{tab_diff_scores_fr}
\end{table}
\begin{table}[h]
\small \center
\rowcolors{1}{white}{gris}
\begin{tabular}{l|ll|ll|ll|ll} 
\hline
&\multicolumn{2}{c|}{C}&\multicolumn{2}{c|}{P}&\multicolumn{2}{c}{OQ}\\
\hline \hline		
&-f&+f&-f&+f&-f&+f\\
\hline	 \hline	
\footnotesize B&.06&.13&.80&.35&.05&.05\\
\footnotesize BS&.08&.16&.69&.31&.05&.05\\
\footnotesize BM&.12&.22&.40&.23&.05&.05\\
\footnotesize BC&.17&.26&{\bf.65}&{\bf.39}&{\bf.11}&{\bf.10}\\
\footnotesize BCSM&{\bf.24}&.{\bf36}&.43&.27&.10&{\bf.10}\\
\hline \hline	
avg. gain&\multicolumn{2}{c|}{\cellcolor[gray]{0.9}+9.2}&\multicolumn{2}{c|}{\cellcolor[gray]{0.9}-28.4}&\multicolumn{2}{c|}{\cellcolor[gray]{0.9}-0.2}\\
\hline	 \hline		
\end{tabular}
\caption{Scores without (-f) and with (+f) fertile translations  (EN$\rightarrow$DE)}
\label{tab_diff_scores_de}
\end{table}

\section{Conclusion and future work}
\label{sec:conclusion}
We have proposed a method based on the compositionality principle which can extract translations pairs from comparable corpora. It is capable of dealing with a largely variety of morphologically constructed  terms and can generate \emph{fertile} translations.
The added value of the fertile translations is clear-cut for English to French translation but not for English to German translation. The English-German lexicon is better without the fertile translations. It seems that the added-value of fertile translations depends on the morphological type of the languages involved in translation. Future work includes the improvement of the identification of morphological variants. The morphological families extracted by the stemming algorithm are too broad for the purpose of translation. For example, the words \emph{desirability} and \emph{desiring} have the same stem but they are too distant semantically to be used to generate  translation variants. We need to restrict the morphological families to a small set of morphological relations (e.g. noun $\leftrightarrow$ relational adjective links). We will also work out a way to rank the candidate translations. Several lines of research are possible : go beyond the target corpora and learn a language model from a larger target corpus, mix compositional translation with a context-based approach, learn part-of-speech patterns translation probabilities from a parallel corpora (e.g. learning that it is more probable that a noun is translated as another noun or as a noun phrase rather than an adverb). A last improvement could be to gather morpheme correspondences from parallel data.

%
%
%

\bibliographystyle{apalike}
\bibliography{biblio}

%
%
%
%
%
%
%

\end{document}